\documentclass[11pt]{article}

\usepackage[preprint]{acl}

\usepackage{times}
\usepackage{latexsym}
\usepackage[T1]{fontenc}
\usepackage[utf8]{inputenc}
\usepackage{microtype}
\usepackage{graphicx}
\usepackage{subcaption}
\usepackage{tikz}
\usetikzlibrary{positioning}
\usepackage{booktabs}
\usepackage{amsmath}
\usepackage{pifont}

\title{A Small Model Asks Better: Fine-Tuning Query Clarification for Retrieval-Augmented Generation}

\author{Lihong He, Lucian Popa, Marina Danilevsky \\
  IBM Research, USA \\
  \texttt{lihong.he@ibm.com}  \\
  }

\begin{document}
\maketitle

\begin{abstract}
Query clarification is essential for retrieval-augmented generation (RAG) systems
that face ambiguous user queries. 
We consider multi-turn grounded conversational scenarios where  
a user query is either ill-formed or ambiguous in the context of the prior dialog or, 
more interestingly, it may be well-formed but when posed against retrieved documents there 
are multiple relevant interpretations. The task is to recognize 
both such cases and then generate an appropriate clarification request to the user. 
We present ClariEdge, a new dataset of approximately 3,900 examples 
that include such ambiguous queries along with negative examples that 
do not require clarification. 
We also show that a lightweight fine-tuned LoRA achieves  
96.6\% classification accuracy and outperforms prompted baselines of many larger models on 
both classification and quality of clarification generation. 
Our study shows that hard negatives  
are an essential ingredient 
to separate out queries that are truly ambiguous from queries that 
may look ambiguous but are not (i.e., can be resolved from the context).  
The same LoRA achieves high performance in both pre-retrieval and post-retrieval 
scenarios; thus, it can be used in two places in a RAG pipeline as a check before 
answer generation. 
The ClariEdge dataset is available at \url{https://huggingface.co/datasets/ibm-research/ClariEdge}.
\end{abstract}

\section{Introduction}
\label{sec:introduction}

Ambiguity in user queries can come in two distinct forms. The first is the
\emph{ill-formed} query, where the user mentions an entity or attribute that the
conversation has not narrowed down to a single choice. Figure~\ref{fig:conv-illformed}
shows such a case: after the assistant has surfaced two credit cards, the user
asks ``What are the potential rewards of this credit card?'' The system
must decide whether to attempt an answer under uncertainty or ask which card the
user means; a useful clarification lists the candidates already on the table
(a pattern we call \emph{hedging over list}).

\begin{figure*}[t]
  \centering
  \tikzset{
    bubble/.style={draw=black!50, rounded corners=2pt, align=left, inner sep=3.5pt, font=\footnotesize},
    user/.style={bubble, fill=gray!10},
    asst/.style={bubble, fill=blue!4},
    docs/.style={bubble, draw=black!40, dashed, fill=yellow!8, font=\scriptsize\itshape},
    ambig/.style={bubble, draw=red!75!black, line width=0.7pt, fill=red!6},
    clarif/.style={bubble, draw=blue!70!black, line width=0.7pt, fill=blue!10},
  }
  \begin{subfigure}[t]{0.48\textwidth}
    \centering
    \begin{tikzpicture}
      \node[user, text width=0.92\linewidth] (u1) {\textbf{U:} I'm applying for a credit card for personal travel and business expenses. What would you suggest?};
      \node[asst, text width=0.92\linewidth, below=1mm of u1] (a1) {\textbf{A:} Two strong options: (1)~\emph{Gold Card} --- travel rewards and lounge access; (2)~\emph{Business Card} --- cashback on office expenses.};
      \node[ambig, text width=0.92\linewidth, below=1mm of a1] (u2) {\textbf{U:} What are the rewards of \textbf{this credit card}?};
      \node[clarif, text width=0.92\linewidth, below=1mm of u2] (a2) {\textbf{A:} Which credit card are you asking about: the Gold Card or the Business Card?};
    \end{tikzpicture}
    \caption{Ill-formed query: ``this credit card'' has multiple referents in the conversation. Clarification lists the candidates (\emph{hedging over list}).}
    \label{fig:conv-illformed}
  \end{subfigure}\hfill
  \begin{subfigure}[t]{0.48\textwidth}
    \centering
    \begin{tikzpicture}
      \node[ambig, text width=0.92\linewidth] (u1) {\textbf{U:} Is \textbf{the language} that descended from Common Brittonic still spoken today, or has it become extinct?};
      \node[docs, text width=0.92\linewidth, below=1mm of u1] (d) {Retrieved: passages on Welsh, Breton, and Cornish as descendants of Common Brittonic.};
      \node[clarif, text width=0.92\linewidth, below=1mm of d] (a1) {\textbf{A:} Several languages descended from Common Brittonic: \emph{Welsh}, still widely spoken in Wales; \emph{Breton}, still spoken in Brittany; and \emph{Cornish}, extinct in the 19th century but since revitalized. Which one are you referring to?};
    \end{tikzpicture}
    \caption{Well-formed query: user thinks there is only one choice, but the retrieved corpus shows several. Clarification pairs each option with a brief grounded answer (\emph{hedging with answers}).}
    \label{fig:conv-wellformed}
  \end{subfigure}
  \caption{Ambiguity forms in conversational RAG. \textcolor{red}{Ambiguous user query} and \textcolor{blue}{clarification response} are highlighted.}
  \label{fig:conv_clarify}
\end{figure*}

The second form arises when the query is \emph{well-formed in
isolation} but becomes ambiguous once grounded in a set of retrieved
documents. In Figure~\ref{fig:conv-wellformed}, the user asks whether
``the language that descended from Common Brittonic'' is still spoken, a question that sounds well-defined - but the retrieved passages surface several possible descendants (Welsh, Breton, Cornish). In retrieval-augmented generation (RAG) \citep{rosenthal2026mtragun}, this is a common occurrence, and a useful clarification should both flag the ambiguity and also
enumerate the grounded options with a brief answer for each (\emph{hedging with
answers}). The \emph{quality} of clarification, that is, 
naming the right options and choosing the pattern of communication
(see Table~\ref{tab:clar-patterns} for details)
matters as much as the
decision of whether to clarify at all. Both use cases -  ambiguity before retrieval, and after - can occur in the same deployment. 
Therefore we need a unified clarification functionality that handles both, usable before retrieval, based only on the conversation 
\citep{cao2025icr, sahay2025ask}, and after retrieval, when documents become available
\citep{kim2023toc}.

Existing benchmarks address only one slice of this picture. CLAMBER
\citep{zhang2024clamber} evaluates ambiguity detection on standalone questions (no conversational context, no retrieved documents) and finds that
chain-of-thought and few-shot prompting yield only marginal gains over a base
classifier. AmbigQA \citep{min2020ambigqa} grounds ambiguity in a corpus, but
treats a question as ambiguous whenever retrieval surfaces multiple atomic
factual answers, even when they are part of the same larger answer. This is a 
notion that current conversational LLMs largely subsume by responding with 
a composite answer that includes the atomic facts: for example, ``Who took control of the Italian government in 1922?''%
\footnote{This question, along with many similar ones, comes from the AmbigQA dataset and is considered ambiguous there.} 
yields ``Benito Mussolini'' and ``the Fascists'' in one response -- 
so the AmbigQA ambiguity rarely surfaces as a clarification need. 
Neither CLAMBER nor AmbigQA capture the joint conversation-plus-corpus setting 
that real RAG deployments face. 

\paragraph{Problem Statement.} Given a multi-turn conversation between a user and an
AI assistant, together with optionally retrieved documents for the last user turn, our goal is to determine
whether the user's last turn is ambiguous, that is, whether it admits either no
clear interpretation or multiple valid interpretations. When it is ambiguous,
the system should generate a clarification request that elicits additional information from the user.

Towards addressing this problem, we make the following specific contributions:

\textbf{1. ClariEdge: Clarification Grounded Dataset with Hard Negative Edge Cases.}
We release ClariEdge, a dataset of approximately 3,900 examples with multi-turn
conversations, retrieved documents, and explicit hard negatives, that is,
queries that look ambiguous but are resolvable from context via coreference
resolution or query rewriting. The dataset is built
through direct human annotation and synthetic LLM generation followed by human
validation.

\textbf{2. Lightweight Fine-Tuning Substantially Outperforms Prompting.}
Against seven prompted baselines that include substantially larger models, our
lightweight LoRA achieves the highest classification accuracy
and the best clarification-generation quality on the queries it
correctly flags.
Whereas prompted baselines tend to either over-clarify or under-clarify, the
fine-tuned model handles both directions reliably.
An ablation study further
confirms that the hard negatives present in ClariEdge are essential for query clarification, since removing
them causes a sharp drop in accuracy on boundary cases.

\begin{figure}[t]
  \centering
  \includegraphics[width=.95\linewidth]{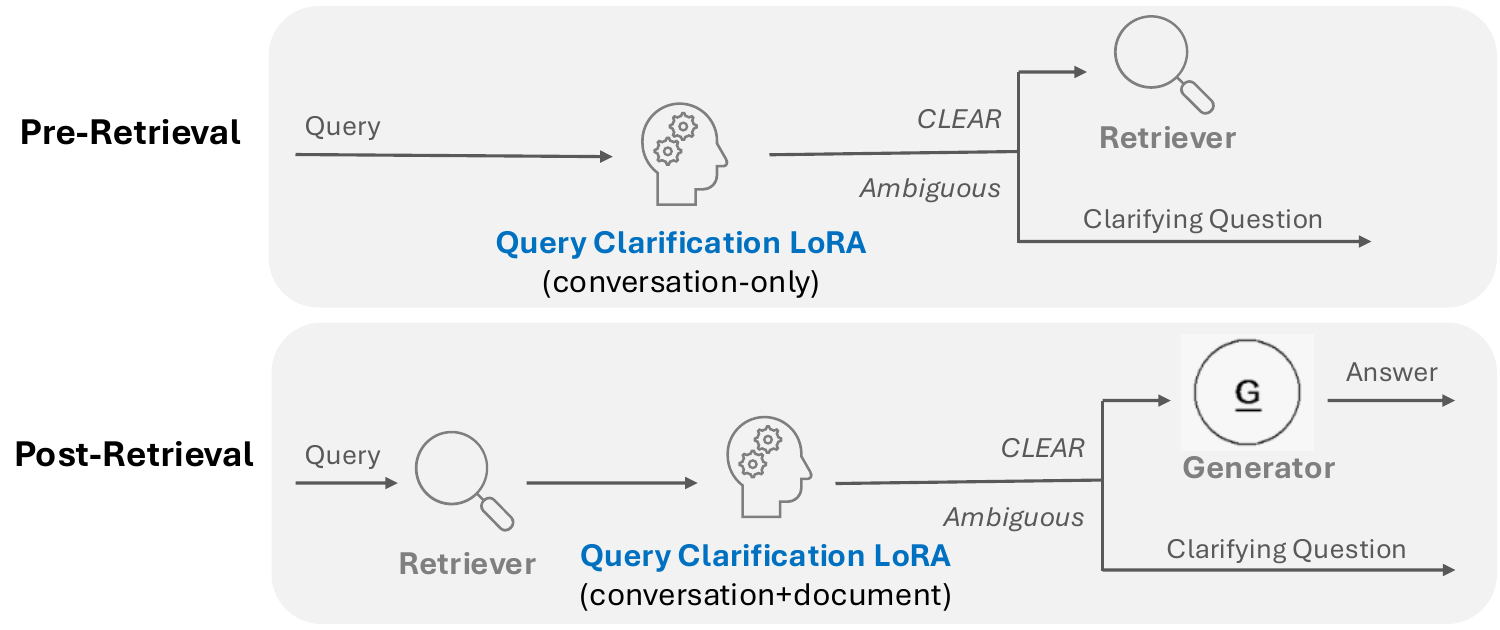}
  \caption{Deployment in a RAG pipeline.}
   \label{fig:rag-deployment}
\vspace{-0.1in}
\end{figure}

\textbf{3. Multi-Setting Handling in a Single Model.} Unlike approaches that rely on
separate models or dynamic strategy selection \citep{sahay2025ask}, one LoRA model handles
both pre-retrieval (conversation-only) and post-retrieval (conversation plus
documents) scenarios, with classification accuracy ahead of larger prompted models.
As shown in Figure~\ref{fig:rag-deployment}, the same LoRA is invoked once on the
conversation alone (gating retrieval) and once on the conversation plus retrieved
documents (gating answer generation), asking a clarifying question whenever either
invocation flags ambiguity. 

\section{Related Work}
\label{sec:related-work}

Ambiguity and clarification have been studied across many information-seeking
settings: open-domain factual QA (AmbigQA, \citealp{min2020ambigqa}; ASQA,
\citealp{stelmakh2022asqa}; CAmbigNQ, \citealp{lee2023asking}; CondAmbigQA,
\citealp{li2025condambigqa}), standalone questions (CLAMBER,
\citealp{zhang2024clamber}), conversational QA (Abg-CoQA,
\citealp{guo2021abgcoqa}; PACIFIC, \citealp{deng2022pacific}; CANARD,
\citealp{elgohary2019can}), and conversational search (Qulac,
\citealp{aliannejadi2019qulac}; ClariQ, \citealp{aliannejadi2020convai3};
MIMICS, \citealp{zamani2020mimics}). Table~\ref{tab:dataset-comparison} 
compares our dataset against four of these
benchmarks. On the modeling side, LLMs gain only marginal benefits from
chain-of-thought or few-shot prompting on ambiguity identification
\citep{zhang2024clamber, deng2023proactive}, motivating work that exploits
models' own uncertainty signals \citep{kim2024apa, testoni2024asking} or
resolves ambiguity via retrieval-grounded disambiguation without asking the
user \citep{kim2023toc}.

On the method side, prior systems specialize clarification along three axes:
\emph{when} to clarify, using LLM uncertainty or calibration
\citep{kuhn2023clam, zhang2023clarify, cole2023selectively}; \emph{how} to
ask, via facet-driven or reasoning-trace supervision
\citep{sekulic2021facet, andukuri2024stargate, zhang2025future,
rahmani2024clarifying}; and \emph{which} deployment setting to target
\citep{sahay2025ask, chen2024style, cao2025icr, qin2025pamis}. These systems
typically rely on prompting or uncertainty rather than task-specific
supervision, and each remains tied to the prompts, models, or iterations of
its setting.

In contrast, we show that a single LoRA fine-tuned with hard negatives covers
both pre- and post-retrieval, draws an explicit boundary between genuine
ambiguity and context-resolvable references, and outperforms substantially larger
prompted baselines. This indicates that training-data quality dominates raw model
scale on this task.

\begin{table}[t]
  \centering
  \footnotesize
  \setlength{\tabcolsep}{3pt}
  \begin{tabular}{l l c c c r}
    \toprule
    Dataset & \shortstack{Ambiguity\\type} & \shortstack{Multi-\\turn} & \shortstack{Retri-\\eval} & \shortstack{Hard\\negatives} & Size \\
    \midrule
    AmbigQA   & answer-set    & \ding{55} & \ding{51} & \ding{55} & 14k  \\
    CLAMBER   & typology      & \ding{55}   & \ding{55} & \ding{55} & 12k  \\
    ClariQ    & facet         & \ding{55} & \ding{51} & \ding{55} & 237 \\
    Abg-CoQA  & referential   & \ding{51} & \ding{51} & \ding{55} & 8.6k   \\
    \midrule
    \textbf{ClariEdge} & entity-option & \ding{51} & \ding{51} & \ding{51} & 3.9k \\
    \bottomrule
  \end{tabular}
  \caption{Comparison with prior clarification datasets.}
  \label{tab:dataset-comparison}
\end{table}

\section{ClariEdge Dataset}
\label{sec:dataset}

Effective query clarification requires training data that captures the nuanced
decision boundaries between clarification-worthy and clear queries. This section describes our  ClariEdge dataset and its creation. 
Conversations span the same four domains as MTRAG \citep{katsis2025mtrag}: ClapNQ (Wikipedia), Cloud (technical documentation), FiQA (finance), and Govt (government).

Table~\ref{tab:dataset-comparison} compares our ClariEdge dataset against four widely used
clarification benchmarks, each covering a complementary slice:
AmbigQA \citep{min2020ambigqa} targets \emph{answer-set} ambiguity in open-domain factual QA; CLAMBER \citep{zhang2024clamber} introduces a \emph{Epistemic Misalignment / Linguistic Ambiguity / Aleatoric Output}
typology for LLM evaluation on standalone questions; ClariQ \citep{aliannejadi2020convai3} captures
web-search \emph{facets}; and Abg-CoQA \citep{guo2021abgcoqa} addresses
\emph{referential} ambiguity in conversational QA. None of these combines
multi-turn conversation, retrieved documents, and explicit hard negatives in a
single resource. Our dataset targets \emph{entity-option} ambiguity, where the
user query may be referring to one of several candidate entities surfaced by either the conversation or the retrieved documents, and is the first to include systematic hard
negative edge cases that look ambiguous but are resolvable from context.

\subsection{Data Categories and Statistics}
\label{sec:data-categories}

\begin{table}[t]
  \centering
  \small
  \begin{tabular}{llrrr}
    \toprule
    Category & Type & Train & Test & All \\
    \midrule
    Ambiguous & positive & 967 & 104 & 1{,}071 \\
    CLEAR-random & easy negative & 1{,}800 & 200 & 2{,}000 \\
    CLEAR-hard & hard negative & 742 & 79 & 821 \\
    \midrule
    \textbf{Total} & & \textbf{3{,}509} & \textbf{383} & \textbf{3{,}892} \\
    \bottomrule
  \end{tabular}
  \caption{ClariEdge dataset categories and statistics.}
  \label{tab:dataset-stats}
\end{table}

\begin{table*}[t]
  \centering
  \small
  \begin{tabular}{p{2.2cm}p{3.5cm}p{4.3cm}p{4.5cm}}
    \toprule
    Category & User Query & Context & Why This Category \\
    \midrule
    Ambiguous & ``How did the fast food chain establish a significant presence in the US?'' & Documents describe many fast food chains such as In-N-Out Burger, Jack in the Box & Multiple valid interpretations for fast food chain, the user must specify which one \\
    CLEAR-random & ``What is the capital of France?'' & Documents contain a direct answer & Unambiguous query with a single clear interpretation \\
    CLEAR-hard & ``When are the services of this division available?'' & Prior turn: ``I work in the Northeast division'' & Appears ambiguous (``this division'') but resolvable from prior context \\
    \bottomrule
  \end{tabular}
  \caption{Examples for each data category.}
  \label{tab:data-examples}
\end{table*}

Our dataset comprises three categories: (1) \textbf{Ambiguous} (positive examples),
where user queries genuinely require clarification due to multiple plausible
interpretations; (2) \textbf{CLEAR-random} (easy negative examples), which are queries
that do not require clarification and are straightforwardly clear; and (3) \textbf{CLEAR-hard} 
(hard negative examples), which are queries that appear superficially ambiguous
but can be resolved from conversation context. Table~\ref{tab:dataset-stats} presents
the data size and Table~\ref{tab:data-examples} illustrates some examples for the three categories.

The distinction between CLEAR-hard and CLEAR-random is critical:
hard negatives appear ambiguous due to referential expressions but
can be resolved from conversation context, while easy negatives are straightforwardly
clear queries. Training with hard negatives teaches the model the precise boundary
between genuinely ambiguous queries and those with resolvable references.

Additional dataset statistics (pre-/post-retrieval split, turn counts,
document counts) are reported in Appendix~\ref{sec:data-distributions-appendix}.

\subsection{Data Creation Process}
\label{sec:data-creation}

\paragraph{Ambiguous examples.} Following the methodology for creating underspecified
questions in MTRAG-UN \citep{rosenthal2026mtragun}, the final user query in each
conversation was designed to admit multiple plausible interpretations. Conversations
were constructed via two parallel approaches: (1) annotators writing ambiguous
questions inspired by entity types and instances drawn from the underlying
corpus; each such question was then stitched, in a subsequent step, as the last
turn onto an existing human-annotated multi-turn conversation, or (2) synthetic
generation of ambiguous questions appended as the final turn to existing
human-created multi-turn conversations, followed by a manual validation step. The latter approach is also the same one used for CLEAR-hard examples (see more details later): the major difference is that here the manual validation decides that these are truly ambiguous questions, while for CLEAR-hard it decides the converse.

For approach (1), relevant passages were added to each conversation using two
retrieval strategies: direct retrieval with the ambiguous question, and query
expansion with a context relevance filter, which yields a richer set of passages
to simulate scenarios where significantly more interpretations exist; these
examples form the post-retrieval portion of the Ambiguous category. Further details on the human annotation guidelines for this approach are given in Appendix~\ref{sec:human-authored-ambig}. For approach
(2), no passages were retrieved: this branch shares the synthetic-then-validate
pipeline used for CLEAR-hard (described below), and produces conversation-only
(pre-retrieval) Ambiguous examples. In both cases, reference clarification
responses were generated using an LLM with pattern-based prompting to include
relevant disambiguation options (see Section~\ref{sec:clar-gen}), followed by
human correction.

\paragraph{CLEAR-random examples.} The easy negative examples in this category were 
constructed by randomly sampling clear, unambiguous user queries from existing 
multi-turn conversations. For some examples, retrieved documents were paired with 
the query to represent clear post-retrieval scenarios; for others, documents were 
deliberately removed to create conversation-only examples where the query remains 
unambiguous without external context.

\paragraph{CLEAR-hard examples.} A key contribution of our dataset is the systematic
construction of hard negatives, which are queries that may appear ambiguous to an LLM but can 
be resolved from conversation context. 
By construction, these examples are for the conversation-only (pre-retrieval) setting: their ambiguity-or-resolvability hinges on linguistic cues in the
dialogue rather than on retrieved passages.
We describe their creation next.

We prompted an LLM
on multi-turn conversations to generate questions targeting
entities discussed in the conversation, where the last user query is intended to be ambiguous. 
In the basic variant, the model was given 793 conversations and asked to produce questions
where a user asks about an entity type with multiple instances occurring in the conversation. In
an advanced variant, 228 longer conversations (>6 turns with multiple entity types)
were processed using chain-of-thought reasoning with self-verification and iterative
refinement. Human annotators then manually reviewed all LLM-generated queries,
classifying each as genuinely ambiguous or resolvable from context. Conversations with 
last user query determined to be truly ambiguous were moved to the Ambiguous category, while 
the remaining ones, where queries appear ambiguous but can be resolved through conversation
context, became hard negatives. 

We collected additional patterns of hard negatives where the synthetic  generation model incorrectly generated clear queries as ambiguous. These included: ``which one'' patterns where the phrasing appears ambiguous but the question is just a comparison across the previously mentioned entities; follow-up questions resolvable through the regular conversation flow; and terse or keyword queries that appear ambiguous but are clear in context. 
Table~\ref{tab:hard-neg-patterns} gives examples of these additional patterns.

In the above process, we used Mistral-Large-3\footnote{\url{https://huggingface.co/mistralai/Mistral-Large-3-675B-Instruct-2512-NVFP4}} for synthetic generation. Even though this model is relatively powerful, we were able to obtain enough CLEAR-hard examples, showing that the model struggles to delineate the boundary between ambiguous and clear. The human manual review was essential for this. Further details on the manual review process, including an illustrative example, are given in Appendix~\ref{sec:manual-review-synthetic}.

\begin{table}[t]
  \centering
  \small
  \begin{tabular}{p{1.5cm}p{2.2cm}p{3.0cm}}
    \toprule
    Pattern & User Query & Why CLEAR \\
    \midrule
    ``Which one'' comparison & ``Which one has the best battery life?'' & Comparing all entities, not asking about one specific \\
    Follow-up question & ``What's the cancellation policy?'' & General question about all discussed options \\
    Terse query & ``Price?'' & No pronoun; asking about all entities, not one specific \\
    \bottomrule
  \end{tabular}
  \caption{Additional hard negative patterns.}
  \label{tab:hard-neg-patterns}
\end{table}

\subsection{Generation of Clarification Response}
\label{sec:clar-gen}

For each ambiguous query in the dataset, we generate a reference clarification
response through a structured pipeline:
(1) identify the ambiguous entity in the last user query, that is, the term or
reference that admits multiple interpretations;
(2) extract candidate entity options from the retrieved document and prior
conversations, together with a short associated answer where applicable;
(3) verify the consistency of each extracted option against the ambiguous entity
and deduplicate them;
(4) generate the clarification response based on the number of distinct options,
following one of three patterns described below;
(5) human annotators review the final clarification response and modify it when
the LLM-generated output is inaccurate or insufficient.

Steps 1--4 are performed by an
LLM (the same Mistral-Large-3), 
followed by human review and correction in step 5.
In step 4, the clarification response follows one of three patterns based on the number 
of distinct options: (1) Hedging with answers (few options, e.g., 2--3): list each option 
with a brief description or answer; (2) Hedging over list (medium number of options, 
e.g., 4--8): enumerate the plausible options without additional explanatory content; 
and (3) Open-domain (many or unbounded options): directly asks the user for disambiguation 
over the type of entity that they may have in mind. 
Appendix~\ref{sec:clar-patterns-appendix} provides some examples of the three patterns.

\subsection{Inter-rater Agreement}
\label{sec:inter-rater}

To measure the annotation reliability, we follow the calculation of
inter-rater agreement in Abg-CoQA~\citep{guo2021abgcoqa}.

For the inter-rater agreement on whether a user query is ambiguous
or not, we randomly select 80 samples from the test set: 20 from CLEAR-hard,
40 from CLEAR-random, and 20 from Ambiguous. We ask two authors to provide
their annotations (CLEAR or Ambiguous) on these 80 examples, and then
compute Cohen's Kappa. The two annotators disagree on only 4 data points,
all of which are sampled from the CLEAR-hard category. The resulting
Cohen's Kappa is 0.875 (almost perfect agreement), indicating that the
ambiguity labels are highly reliable.

For the inter-rater agreement on clarification questions, we ask
the same two annotators to independently write a clarification question for
each of the 20 sampled Ambiguous examples (one version is kept as the gold
clarification). We then compute the macro-average F1 score of word overlap
and the BLEU-4 score. The F1 is 0.774 and the BLEU-4 is 0.57. We further
conduct a manual review of the two clarification versions on the 20
samples. The manual inspection and the metrics agree that the two versions
match on the ambiguous entity and its candidate options but differ in
surface phrasing: the F1 of 0.774 captures the shared content while the
moderate BLEU-4 of 0.57 reflects the phrasing variation, indicating
that the gold clarifications are content-faithful yet not unique verbatim
references.

\section{Models}
\label{sec:model-deployment}

This section describes our fine-tuned LoRA models and the baseline models used in the experiments.

\subsection{Fine-Tuned LoRA Models}
\label{sec:lora-models}

We employ Low-Rank Adaptation (LoRA) \citep{hu2022lora} to create
parameter-efficient clarification models, fine-tuned on the 3{,}509 training examples
described in Table~\ref{tab:dataset-stats}. 
This substantially reduces training cost and enables rapid experimentation with different training data compositions.
We fine-tune LoRA on (1) small 3B models from the Granite family
(Granite-4.0-micro\footnote{\url{https://huggingface.co/ibm-granite/granite-4.0-micro}} 
and Granite-4.1-3B\footnote{\url{https://huggingface.co/ibm-granite/granite-4.1-3b}}), 
targeted at cost-effective deployment; and 
(2) GPT-OSS-20B, a larger open-source model included for comparison.

The two small Granite models serve
as our primary models. This choice reflects several considerations. First,
their size-to-performance trade-off provides sufficient capacity while
maintaining deployment efficiency, in line with recent evidence that
lightweight models can be competitive on well-defined tasks
\citep{nandi2026scaling}. Second, both are amenable to LoRA
adaptation with a modest training set. Finally, using the same base
architectures for both fine-tuned and prompted configurations allows us to
isolate the effect of fine-tuning from any difference in model capacity.

Training configuration and input/output format for the LoRA models are
described in Appendix~\ref{sec:lora-details}.

\subsection{Prompted Baseline Models}
\label{sec:baselines}

To quantify the value of fine-tuning, we compare our LoRA models against seven prompted
baselines evaluated without any task-specific training. The baselines span proprietary
frontier, open-weights, and lightweight model families:
\textbf{GPT-5.6 Sol} and \textbf{GPT-4o} (frontier, proprietary);
\textbf{GPT-OSS-120B} (large, open-source);
\textbf{GPT-OSS-20B} (medium, open-source);
\textbf{Granite-4.1-30B} (medium, open-source); and
\textbf{Granite-4.1-3B} and \textbf{Granite-4.0-micro} (lightweight, open-source).

All prompted baselines receive identical chat messages consisting of a system message specifying the clarification rules and a user message constructed from each test record.
The system message defines the task, grounding requirements, and  describes the patterns that the generated clarifications should follow. The user message supplies the conversation and retrieved documents (if any). 
The full text of both messages is provided in Appendix~\ref{sec:baseline-prompts}.

\section{Experiments and Analysis}
\label{sec:experiments}

\begin{table*}[t]
  \centering
  \footnotesize
  \setlength{\tabcolsep}{3pt}
  \begin{tabular}{l cccc cccc ccc}
    \toprule
    & \multicolumn{4}{c}{\emph{Full Test Set}} & \multicolumn{4}{c}{\emph{Pre-Retrieval}} & \multicolumn{3}{c}{\emph{Post-Retrieval}} \\
    \cmidrule(lr){2-5} \cmidrule(lr){6-9} \cmidrule(lr){10-12}
    Model & Ambig. & C-rand. & C-hard & Overall & Ambig. & C-rand. & C-hard & Overall & Ambig. & C-rand. & Overall \\
    \midrule
    Granite-4.1-3B LoRA    & 91.3\%          & 99.5\%          & 96.2\%          & 96.6\%          & 84.5\%          & 98.0\%           & 96.2\%          & 93.0\%          & 100.0\%          & 100.0\%          & 100.0\%          \\
    Granite-4.0-micro LoRA & 84.6\%          & 98.5\%          & 87.3\%          & 92.4\%          & 72.4\%          & 98.0\%           & 87.3\%          & 85.6\%          & 100.0\%          & 98.7\%           & 99.0\%           \\
    GPT-OSS-20B LoRA       & 76.9\%          & 98.5\%          & 86.1\%          & 90.1\%          & 62.1\%          & 100.0\%          & 86.1\%          & 82.4\%          & 95.7\%           & 98.0\%           & 97.4\%           \\
    \midrule
    GPT-4o            & 48.1\% & 90.5\% & 92.4\% & 79.4\% & 60.3\% & 88.0\% & 92.4\% & 81.3\% & 32.6\% & 91.3\% & 77.6\% \\
    GPT-5.6 Sol       & 93.3\% & 72.5\% & 48.1\% & 73.1\% & 94.8\% & 70.0\% & 48.1\% & 68.4\% & 91.3\% & 73.3\% & 77.6\% \\
    GPT-OSS-120B      & 80.8\% & 83.0\% & 65.8\% & 78.9\% & 86.2\% & 78.0\% & 65.8\% & 75.4\% & 73.9\% & 84.7\% & 82.1\% \\
    GPT-OSS-20B       & 90.4\% & 72.5\% & 58.2\% & 74.4\% & 93.1\%          & 72.0\% & 58.2\% & 72.7\% & 87.0\% & 72.7\% & 76.0\% \\
    Granite-4.1-30B   & 78.8\% & 80.0\% & 63.3\% & 76.2\% & 87.9\% & 78.0\% & 63.3\% & 74.9\% & 67.4\% & 80.7\% & 77.6\% \\
    Granite-4.1-3B    & 26.9\% & 88.0\% & 55.7\% & 64.8\% & 34.5\% & 94.0\% & 55.7\% & 59.4\% & 17.4\% & 86.0\% & 69.9\% \\
    Granite-4.0-micro & 25.0\% & 85.0\% & 63.3\% & 64.2\% & 31.0\% & 90.0\% & 63.3\% & 60.4\% & 17.4\% & 83.3\% & 67.9\% \\
    \bottomrule
  \end{tabular}
  \caption{Classification accuracy. Ambig.=Ambiguous, C-rand.=CLEAR-random, C-hard=CLEAR-hard.}
  \label{tab:classification}
\end{table*}

In this section, we describe the evaluation metrics and then give  the main experimental results.

\subsection{Evaluation Metrics}
\label{sec:eval-metrics}

We assess model performance along two dimensions: (1) \textbf{classification
accuracy}, measuring whether the model correctly determines if a query is ambiguous or not; and (2) \textbf{generation quality}, evaluating the
clarification response produced for correctly identified ambiguous queries.

Prior work has used BLEU \citep{guo2021abgcoqa} and BERTScore
\citep{zhang2024clamber} to evaluate generation against a reference answer,
but neither captures whether the clarification presents correct
disambiguation options for the ambiguous entity. We therefore design
entity-option-centric metrics in the spirit of AmbigQA
\citep{min2020ambigqa}, differing in that we evaluate over entity options
rather than answer interpretations and accommodate open-domain
clarifications where options cannot be explicitly enumerated.
Specifically, we 
use the following metrics: 
\textbf{Precision}: the fraction of options in the generated clarification that are correct; 
  \textbf{Recall}: the fraction of gold reference options
    that are covered by the generated clarification; 
  \textbf{F1}: harmonic mean of precision and recall.
Computing precision and recall on free-form clarification responses requires four auxiliary judgments. We obtain these judgments with four LLM prompts (see Appendix~\ref{sec:metric-prompts}). We also provide examples in Appendix~\ref{sec:eval-examples-appendix} to illustrate the metrics, as well as how we handle open-domain clarifications.

\subsection{Classification Results}
\label{sec:cls-results}

We evaluate classification accuracy in three settings: full test set,
pre-retrieval subset, and post-retrieval 
subset. The results are reported in Table~\ref{tab:classification}.

\subsubsection{Full Test Set}
\label{sec:cls-full}

The classification results on the full test set support two observations, which we discuss in turn.

\paragraph{Fine-tuning beats prompting at all scales.}
The Granite-4.1-3B LoRA reaches 96.6\% overall accuracy, a 17.2 pp gain
over GPT-4o and a 31.8 pp gain over prompted Granite-4.1-3B. Because the
prompted and fine-tuned variants of Granite-4.1-3B share the same
architecture, this improvement is attributable to training rather than to
greater model capacity, which is consistent with \citet{zhang2024clamber}'s
finding that prompting offers limited gains on the task of query
clarification. The Granite-4.0-micro LoRA also achieves strong
performance at 92.4\%, and both Granite LoRAs outperform the larger
GPT-OSS-20B LoRA and all the other prompted baselines, in line with prior
evidence that specialized lightweight models can be competitive on
well-scoped tasks \citep{nandi2026scaling}. Even the weaker GPT-OSS-20B
LoRA (90.1\%) outperforms the strongest prompted baseline GPT-4o
by 10.7 pp overall, indicating that the gap is robust to LoRA backbone choice.

\paragraph{Prompted baselines are systematically imbalanced.} The imbalance is most
pronounced between the Ambiguous and CLEAR-hard categories, the two that hinge on
correctly distinguishing genuine ambiguity from surface-level ambiguity. Prompted
baselines split into two failure modes along this axis: over-clarifiers (such as GPT-5.6 Sol and GPT-OSS-20B), biased toward flagging queries as ambiguous, and under-clarifiers (such as GPT-4o and the smaller Granite models),
biased toward declaring queries clear. Notably, the two frontier proprietary models diverge along this axis, indicating that the imbalance is not attributable to model scale alone.
In contrast, the fine-tuned LoRAs (and especially the Granite two) maintain a good balance, indicating that exposure to hard negatives is
essential for separating genuine ambiguity from resolvable references.
We also conduct an ablation study on hard negatives, 
showing that 
removing them causes a sharp drop in accuracy on boundary cases. The details 
are in Appendix~\ref{sec:ablation}.

\begin{table*}[t]
  \centering
  \footnotesize
  \setlength{\tabcolsep}{3pt}
  \begin{tabular}{l cccc cccc cccc}
    \toprule
    & \multicolumn{4}{c}{\emph{Full Test Set}} & \multicolumn{4}{c}{\emph{Pre-Retrieval}} & \multicolumn{4}{c}{\emph{Post-Retrieval}} \\
    \cmidrule(lr){2-5} \cmidrule(lr){6-9} \cmidrule(lr){10-13}
    Model & Corr. & Prec. & Rec. & F1 & Corr. & Prec. & Rec. & F1 & Corr. & Prec. & Rec. & F1 \\
    \midrule
    Granite-4.1-3B LoRA    & 95/104 & 92.7\%          & 85.7\%          & 89.0\%          & 49/58 & 89.3\%          & 80.9\%          & 84.9\%          & 46/46 & 96.3\%          & 90.7\% & 93.4\%          \\
    Granite-4.0-micro LoRA & 88/104 & 91.8\%          & 87.7\%          & 89.7\%          & 42/58 & 87.4\%          & 83.6\%          & 85.5\%          & 46/46 & 95.8\%          & 91.4\% & 93.5\%          \\
    GPT-OSS-20B LoRA       & 80/104 & 89.1\%          & 82.3\%          & 85.6\%          & 36/58 & 83.5\%          & 81.9\%          & 82.7\%          & 44/46 & 93.7\%          & 82.6\%          & 87.8\%          \\
    \midrule
    GPT-4o            & 50/104 & 81.3\%          & 69.8\%          & 75.1\% & 35/58 & 74.4\%          & 71.9\%          & 73.1\%          & 15/46 & 97.8\% & 64.2\% & 77.5\% \\
    GPT-5.6 Sol       & 97/104 & 92.6\%          & 73.4\%          & 81.9\% & 55/58 & 90.1\%          & 80.7\%          & 85.1\%          & 42/46 & 96.0\% & 63.9\% & 76.7\% \\
    GPT-OSS-120B      & 84/104 & 89.0\%          & 68.4\%          & 77.3\% & 50/58 & 83.8\%          & 71.5\%          & 77.2\%          & 34/46 & 96.8\%          & 63.7\% & 76.8\% \\
    GPT-OSS-20B       & 94/104 & 93.3\% & 70.7\%          & 80.5\% & 54/58 & 94.0\% & 81.8\%          & 87.5\% & 40/46 & 92.4\%          & 55.1\% & 69.0\% \\
    Granite-4.1-30B   & 82/104 & 88.7\%          & 87.1\% & 87.9\% & 51/58 & 83.7\%          & 88.9\% & 86.2\%          & 31/46 & 97.2\%          & 84.0\% & 90.1\% \\
    Granite-4.1-3B    & 28/104 & 81.8\%          & 72.2\%          & 76.7\% & 20/58 & 78.2\%          & 74.8\%          & 76.5\%          & 8/46  & 91.7\%          & 65.1\% & 76.1\% \\
    Granite-4.0-micro & 26/104 & 75.0\%          & 45.7\%          & 56.8\% & 18/58 & 72.2\%          & 45.9\%          & 56.1\%          & 8/46  & 81.2\%          & 44.7\% & 57.7\% \\
    \bottomrule
  \end{tabular}
  \caption{Generation quality on examples correctly classified as ambiguous.
    Precision (Prec.) and recall (Rec.) are averaged per example within each setting; 
    F1 is their harmonic mean.
    Corr.=correct classified / total ambiguous.}
  \label{tab:generation}
\end{table*}

\subsubsection{Pre-Retrieval}
\label{sec:cls-pre}

In the pre-retrieval setting, the model must decide whether to clarify based solely
on the multi-turn conversation. This setting is 
particularly challenging because the model cannot leverage document evidence
to disambiguate queries and to surface the interpretations.

The classification accuracy on the pre-retrieval subset shows that all three
fine-tuned LoRAs lead on overall accuracy, substantially improving over their
architecture-matched prompted baselines. The class-imbalance pattern observed
on the full test set (\S\ref{sec:cls-full}) is again evident among prompted
models in pre-retrieval: some over-clarify (e.g., GPT-5.6 Sol and GPT-OSS-20B), 
while others under-clarify (e.g., GPT-4o and the smaller
Granite models). In contrast, the Granite-4.1-3B LoRA maintains relatively
balanced performance, though its Ambiguous accuracy is lower than its
CLEAR-hard accuracy.
Pre-retrieval is where linguistic ambiguity is hardest to disentangle:
CLEAR-hard examples by construction (see Section~\ref{sec:data-creation}) are
for the conversation-only setting, and present an inherently harder setup where
the model must rely purely on linguistic cues to separate Ambiguous from cases
that are resolvable in the dialogue.

\subsubsection{Post-Retrieval}
\label{sec:cls-post}

In the post-retrieval setting, the model has access to the conversation together
with documents returned by retriever. The CLEAR-hard category is
not represented in this subset; as we mentioned, they target 
conversational context rather than documents. 

From the post-retrieval results in Table~\ref{tab:classification}, the
fine-tuned LoRAs reach near-saturation overall classification accuracy: the
Granite-4.1-3B LoRA hits 100.0\% and the Granite-4.0-micro LoRA hits 99.0\%.
Both are far ahead of the strongest prompted baseline (GPT-OSS-120B at
82.1\%). The prompted baselines continue to exhibit the class-imbalance
patterns observed in earlier settings.
These results indicate that, once 
retrieved documents are available, the remaining difficulty lies in correctly interpreting 
the retrieval signal. Fine-tuning acquires this skill reliably, whereas prompting does not.

\subsection{Generation Quality Results}
\label{sec:gen-results}

We assess clarification generation quality across the same three settings 
using the entity-option-centric metrics.
A clarification only matters when the model also correctly flags the underlying
query as ambiguous: a high F1 on a small set of correctly-flagged queries is
weaker than the same F1 on a larger set. We therefore report both the count of
correctly-flagged examples and the F1 of the resulting clarifications in
Table~\ref{tab:generation}.

\subsubsection{Full Test Set}
\label{sec:gen-full}

On the full test set, both Granite LoRAs lead on F1: the 4.0-micro LoRA at 89.7\% and the 4.1-3B LoRA at 89.0\%. 
The prompted baselines GPT-5.6 Sol and GPT-OSS-20B post similar high precision but
substantially lower recall, indicating that without task-specific training models
miss plausible alternatives grounded in
the conversation or retrieved documents.
The gain from fine-tuning is especially pronounced for
lightweight models: prompted Granite-4.0-micro reaches only 56.8\% F1, whereas
its LoRA counterpart attains 89.7\%.

\subsubsection{Pre-Retrieval}
\label{sec:gen-pre}

On the pre-retrieval subset, generation quality results are more competitive
across models. Both Granite LoRAs achieve an F1 around 85\%, close to prompted GPT-5.6 Sol,
GPT-OSS-20B and Granite-4.1-30B.
However, as seen in \S\ref{sec:cls-pre}, these three prompted models have much lower classification accuracy (they tend to over-clarify).
The LoRA's operational advantage 
is that it produces high-quality clarifications
while also reliably identifying \emph{which} queries require them, avoiding the
inflation of generation volume that accompanies over-clarification in prompted
baselines.

\subsubsection{Post-Retrieval}
\label{sec:gen-post}

On the post-retrieval subset, the two Granite LoRAs attain the highest F1,
combining high precision with the highest recall. Among prompted models, several
baselines (such as GPT-4o and GPT-OSS-120B) achieve
strong precision but comparatively low recall. This asymmetry suggests that, when
grounded by retrieved documents, prompted models produce confident but narrow
clarifications: they correctly select some document-supported options but frequently
omit others that are equally valid. The two Granite LoRAs' simultaneous gains in precision
and recall indicate that fine-tuning is effective not only at identifying \emph{when}
to clarify but also at enumerating \emph{which} options to present.

\subsection{Out-of-Domain (OOD) Evaluation}
\label{sec:ood}

To assess whether the LoRA's advantage generalizes beyond the four training
domains (ClapNQ, Cloud, FiQA, Govt), we construct an OOD evaluation set in
two new domains drawn from
MTRAG-UN~\citep{rosenthal2026mtragun}: \emph{Banking} (U.S.\ bank services)
and \emph{Telco} (telecommunications providers).
The Granite-4.0-micro LoRA maintains its lead over prompted baselines on
both classification and generation quality. This suggests that the
clarification skill acquired via ClariEdge is task-general rather than
domain-specific. 
Appendix~\ref{sec:ood-appendix} gives the OOD evaluation details.

\section{Conclusion}
\label{sec:conclusion}

In this work, we introduce ClariEdge, a new dataset for clarifying ambiguous questions in RAG settings, and show that LoRA fine-tuning on a small model is more beneficial than prompting larger LLMs. Hard negatives are an essential
ingredient, suggesting that teaching a model what not to do is as important as
teaching it what to do. The resulting LoRA works well in both pre-retrieval and post-retrieval settings, although the pre-retrieval regime is a more challenging one.

\section*{Limitations}

Despite the strong empirical results reported above, our study has several limitations
that circumscribe its scope and suggest directions for future work.

\paragraph{Single-turn clarification.} Our model generates a single clarifying question
rather than engaging in multi-turn clarification dialogues. Practical deployments may
benefit from iterative clarification strategies that progressively narrow the user's
intent across multiple exchanges. Extending our approach
to multi-turn clarification remains an open direction.

\paragraph{English-only evaluation.} Both the dataset and evaluation are restricted to
English. The cross-lingual transferability of our method remains unexplored.

\section*{Acknowledgments}
We would like to thank our annotators for their high-quality work in generating and evaluating this dataset.

\bibliography{custom,anthology-1,anthology-2}

\appendix

\begin{figure*}[t]
  \centering
  \begin{subfigure}{0.32\linewidth}
    \centering
    \includegraphics[width=\linewidth,height=0.65\linewidth,keepaspectratio]{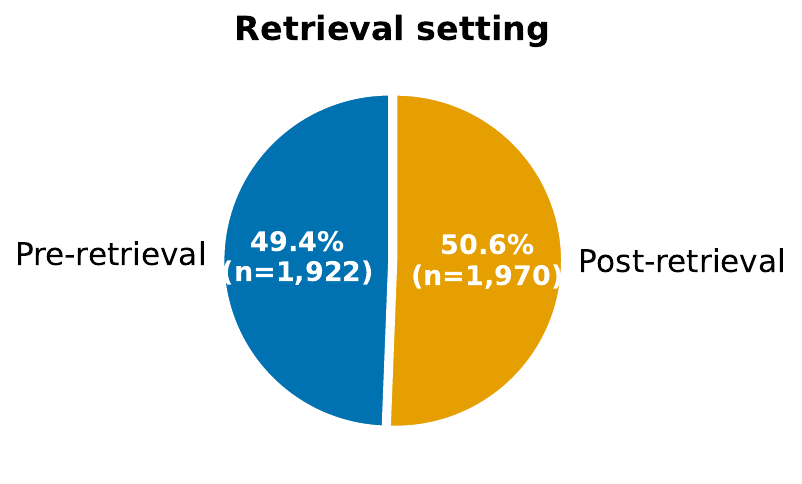}
    \caption{Pre- vs.\ post-retrieval split.}
    \label{fig:pie-retrieval}
  \end{subfigure}\hfill
  \begin{subfigure}{0.32\linewidth}
    \centering
    \includegraphics[width=\linewidth,height=0.65\linewidth,keepaspectratio]{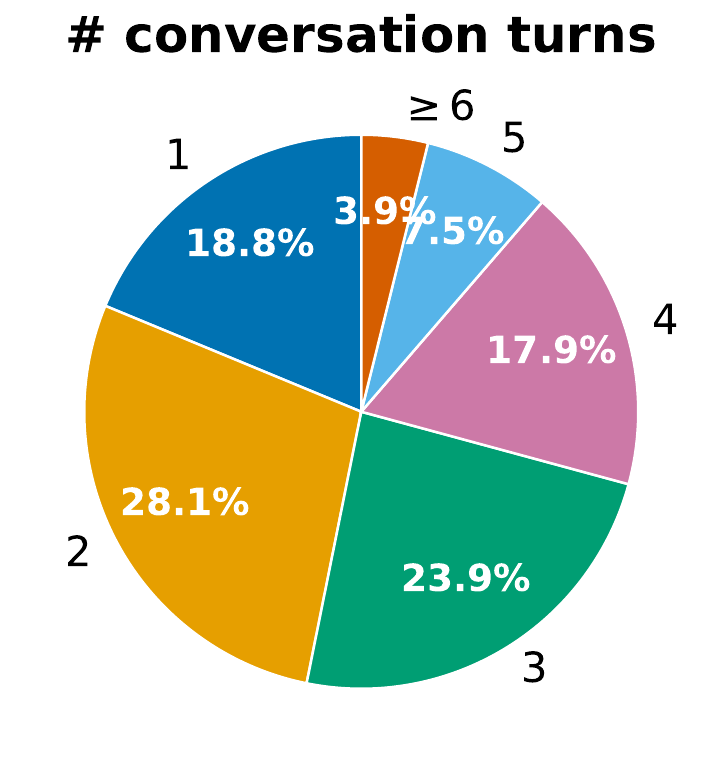}
    \caption{Number of conversation turns.}
    \label{fig:pie-turns}
  \end{subfigure}\hfill
  \begin{subfigure}{0.32\linewidth}
    \centering
    \includegraphics[width=\linewidth,height=0.65\linewidth,keepaspectratio]{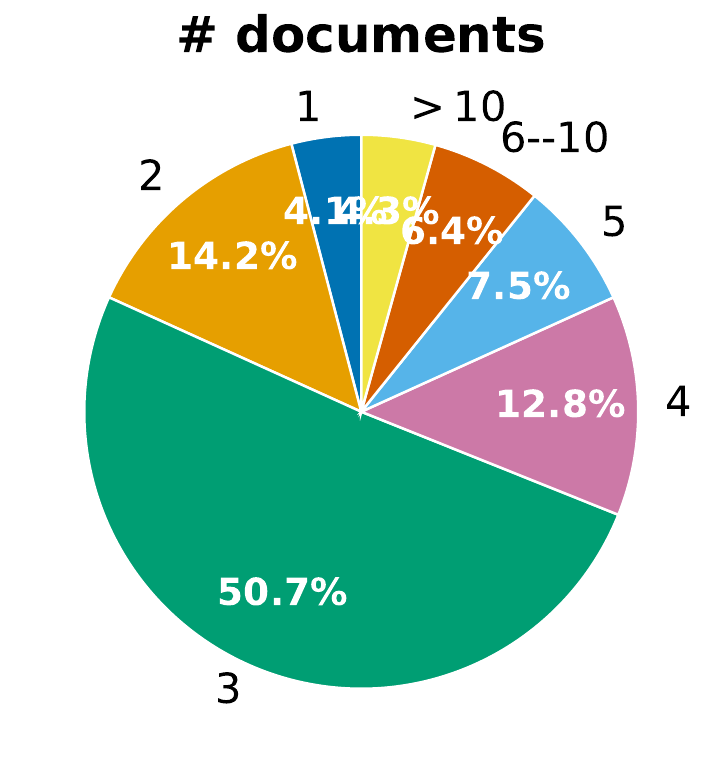}
    \caption{Number of documents (post-retrieval).}
    \label{fig:pie-ndocs}
  \end{subfigure}
  \caption{Data split in pre-/post-retrieval and distribution of conversation
    turns and retrieved document counts.}
  \label{fig:data-distributions}
\end{figure*}

\section{Dataset Distributions}
\label{sec:data-distributions-appendix}

In our dataset, each example is tagged by retrieval setting:
\emph{pre-retrieval} examples expose the model to the conversation only,
whereas \emph{post-retrieval} examples additionally include the documents
returned by the retrieval stage. Figure~\ref{fig:pie-retrieval} shows that
pre-retrieval and post-retrieval examples are represented in roughly equal
proportions. From Figure~\ref{fig:pie-turns}, most conversations are
multi-turn, with only a small fraction consisting of a single turn. From
Figure~\ref{fig:pie-ndocs}, most post-retrieval examples are associated with
at least three retrieved documents.

\section{Human Annotation and Review}
\label{sec:human-ambig-appendix}

This appendix provides additional details on the human annotation and review processes used in creating the dataset (Section~\ref{sec:data-creation}). Section~\ref{sec:human-authored-ambig} focuses on approach (1) for Ambiguous examples, where annotators wrote ambiguous questions inspired by entity types and instances drawn from the underlying corpus, and each such question was then stitched, in a subsequent step, as the last turn onto an existing human-annotated multi-turn conversation. Section~\ref{sec:manual-review-synthetic} focuses on approach (2) for Ambiguous examples and the construction of CLEAR-hard examples, both of which combine LLM-based synthetic generation with manual review (by the authors) to separate truly ambiguous queries from those resolvable from the conversation context. Both of these processes rely on an existing set of general (i.e., with no ambiguity) multi-turn conversations that were created previously by annotators (as described in \cite{katsis2025mtrag}). 

\subsection{Human-Authored Ambiguous Questions}
\label{sec:human-authored-ambig}

Based on the corpus of documents underlying each domain (we used ClapNQ and Gov), we first created a set of \emph{entity types}, each populated with a set of concrete \emph{entity instances} drawn from that corpus. The annotators were then given these entity types and instances, together with instructions to pick an entity type and write an ambiguous question that could be referring to any one of the entity instances of that type. As part of the instructions, the annotators were also given examples of ambiguous questions to illustrate the desired pattern. For example, part of the guidelines read:

\begin{quote}
\emph{If you select the Parks entity, you could ask about the opening hours of a park, as follows: ``What are the park's opening hours?''}
\end{quote}

\noindent where the concrete entity instances available for the Parks entity included: ``Wilder Ranch State Park'', ``Sampson State Park'', ``Leo Carrillo State Park'', ``Mount San Jacinto State Park'', and others.

The annotators were instructed \emph{not} to mention any of the concrete entity instances explicitly in the question, but rather to use them only as inspiration for crafting a question that is ambiguous on purpose. In the example above, the phrase ``the park'' can be referring to any one of the underlying entity instances from the corpus, which is precisely the kind of ambiguity we wish to elicit.

\paragraph{Stitching into multi-turn conversations.} In a subsequent (programmatic) step, each human-authored ambiguous question was stitched as the last turn onto an existing human-annotated multi-turn conversation. Stitching was done in two ways: (a) onto conversations on the same or very similar topic, simulating the case where the new turn is a natural continuation of the dialog (about $75\%$ of the cases), and (b) onto conversations on a different topic, so that the new turn reflects a topic switch by the user while still being an ambiguous question (about $25\%$ of the cases). The latter case adds an additional challenge for the model, reflecting the realistic scenario where users sometimes change topics and the system must still detect ambiguity and react accordingly. 

The resulting conversations went through a careful manual 
validation step (performed by the authors) that included filtering out cases where the addition of the prior conversation context accidentally resolved the intent of the last turn.

\subsection{Manual Review of Synthetically Generated Questions}
\label{sec:manual-review-synthetic}

Approach (2) for Ambiguous examples and the construction of CLEAR-hard examples both rely on LLM-based synthetic generation followed by manual review by the authors of this paper. Concretely, an LLM was prompted to generate a final user query targeting an entity type with multiple instances occurring in the conversation, with the intent that the query be ambiguous (see Section~\ref{sec:data-creation} for the prompting details). However, the synthetic generation step is far from perfect: the model frequently produces questions that, despite mentioning an entity type with several instances, are in fact resolvable from the prior conversation through coreference or normal dialog flow. To address this, we use 
manual review (by the authors).

Figure~\ref{fig:manual-validation-example} illustrates the distinction the manual review must draw. On the left, the user's last turn ``Did he have a significant impact on the movement?'' refers to an entity (a Dadaist artist) for which multiple plausible referents have been introduced in the conversation, and the question cannot be answered without first clarifying which artist; this example is genuinely ambiguous. On the right, the user's last turn ``What are their capabilities?'' superficially looks ambiguous because of the pronoun ``their'', but standard coreference resolution against the immediately preceding agent turn unambiguously binds it to ``bots''; this example is therefore resolvable, and is kept as a CLEAR-hard example rather than as an Ambiguous example.

\begin{figure*}[t]
  \centering
  \tikzset{
    bubble/.style={draw=black!50, rounded corners=2pt, align=left, inner sep=3.5pt, font=\footnotesize},
    user/.style={bubble, fill=gray!10},
    asst/.style={bubble, fill=blue!4},
    ambig/.style={bubble, draw=red!75!black, line width=0.7pt, fill=red!6},
    resolv/.style={bubble, draw=green!55!black, line width=0.7pt, fill=green!6},
  }
  \begin{subfigure}[t]{0.48\textwidth}
    \centering
    \begin{tikzpicture}
      \node[user, text width=0.92\linewidth] (u1) {\textbf{U:} dadaists};
      \node[asst, text width=0.92\linewidth, below=1mm of u1] (a1) {\textbf{A:} The Dadaists were an art movement in the early 20th century, around the time of World War I. They had centers in Z\"urich, Switzerland, New York, and Paris. The artists rejected logic, reason, \ldots};
      \node[ambig, text width=0.92\linewidth, below=1mm of a1] (u2) {\textbf{U:} Did \textbf{he} have a significant impact on \textbf{the movement}?};
    \end{tikzpicture}
    \caption{\textbf{Truly ambiguous} (\ding{51}~kept as Ambiguous): ``he'' has multiple plausible Dadaist referents introduced earlier; cannot be resolved from context.}
    \label{fig:manual-validation-ambig}
  \end{subfigure}\hfill
  \begin{subfigure}[t]{0.48\textwidth}
    \centering
    \begin{tikzpicture}
      \node[user, text width=0.92\linewidth] (u0) {\textbf{U:} \ldots};
      \node[asst, text width=0.92\linewidth, below=1mm of u0] (a1) {\textbf{A:} bots are computer programs that automatically do certain actions on a network. They are programmed to do things at scale \ldots};
      \node[resolv, text width=0.92\linewidth, below=1mm of a1] (u1) {\textbf{U:} What are \textbf{their} capabilities?};
    \end{tikzpicture}
    \caption{\textbf{Resolvable} (\ding{55}~kept as CLEAR-hard): ``their'' is bound by coreference to ``bots'' in the immediately preceding turn; clear interpretation.}
    \label{fig:manual-validation-resolv}
  \end{subfigure}
  \caption{Two synthetically generated last turns that an LLM produced as ``ambiguous''. The manual review step distinguishes between genuinely ambiguous queries (left) and queries that are resolvable from the conversation context, e.g., via coreference (right). The latter become CLEAR-hard examples.}
  \label{fig:manual-validation-example}
\end{figure*}

This boundary, between truly ambiguous queries and queries that merely look ambiguous but can be resolved from the conversational context, is exactly what the synthetic generation models tended to confuse, and is also the boundary that downstream clarification models must learn. The manual review (by the authors) was essential for assigning each generated example to the correct category, and also for surfacing 
various other patterns of false-positive ambiguity (e.g., ``which one'' comparisons, regular follow-up questions, terse keyword queries) in addition to coreference resolution patterns.

\section{Clarification Response Patterns}
\label{sec:clar-patterns-appendix}

This appendix provides representative examples of the three clarification response patterns described in Section~\ref{sec:clar-gen}: \emph{Hedging with answers} (2--3 options), \emph{Hedging over list} (4--8 options), and \emph{Open-domain} (many or unbounded options). Table~\ref{tab:clar-patterns} illustrates each pattern with a user query, the conversational or document context, and the corresponding clarification response.

\begin{table*}[t]
  \centering
  \small
  \begin{tabular}{p{1.2cm}p{1cm}p{2.5cm}p{3.5cm}p{5.5cm}}
    \toprule
    Pattern & \# Opt. & User Query & Context & Clarification Response \\
    \midrule
    Hedging with answers & 2--3 & ``In which country is this language officially spoken?'' & Docs: introduce languages such as Spanish, Italian, and Xhosa & ``There are many languages you could be referring to, such as: 1) Spanish, which is officially spoken in Spain, 2) Italian, which is officially spoken in Italy, Switzerland, San Marino, Vatican City, and western Istria (in Slovenia and Croatia), and 3) Xhosa, which is officially spoken in South Africa and Zimbabwe. Which one are you talking about?'' \\
    Hedging over list & 4--8 & ``How did the fast food chain establish a significant presence in the US?'' & Docs: describe fast food chains such as In-N-Out Burger, Jack in the Box, Big Boy Restaurants, Chipotle Mexican Grill & ``There are many fast food chains you could be talking about, like In-N-Out Burger, Jack in the Box, Big Boy Restaurants, or Chipotle Mexican Grill. Which one are you referring to?'' \\
    Open-domain & $>$8 & ``Tell me about the American poet's work'' & Multiple American poets discussed in conversation & ``Which American poet are you referring to?'' \\
    \bottomrule
  \end{tabular}
  \caption{Clarification response patterns with examples.}
  \label{tab:clar-patterns}
\end{table*}

\section{LoRA Implementation Details}
\label{sec:lora-details}

\paragraph{Training configuration.}
We fine-tune LoRA models with rank 32 and scaling factor $\alpha = 64$,
applied to both attention and MLP layers of the base model. Models are
trained on 1 H100 GPU with learning rate of $1 \times 10^{-5}$ and dropout
of $0.05$. Training runs for up to 15 epochs (minimum 5) with early stopping
based on validation-set performance to mitigate overfitting.

\paragraph{Input and output format.}
Our LoRA models accept flexible inputs: the
conversation history and optional set of retrieved documents for the last user turn. The model adapts its decision based on the available
context: when documents are absent, it relies on the conversation history alone; when
documents are present, it leverages both sources. The output is a JSON object of the form
\texttt{\{"clarification":~"CLEAR" | [clarifying question]\}}. When the query is
unambiguous, the model outputs \texttt{CLEAR}; otherwise, it outputs a clarifying
question (e.g., \emph{``Which credit card are you asking about: the Gold Card or the Business Card?''}).

\section{Prompt for Baseline Models}
\label{sec:baseline-prompts}

This appendix provides the full text of the system and user messages used to evaluate
all seven prompted baselines in Section~\ref{sec:baselines}. All baselines receive the
same templates; only the user-message content varies across test records.

\paragraph{System message.} The system message defines the task, grounding
requirements, and three response patterns corresponding to those used to generate
reference clarifications (Section~\ref{sec:clar-gen}):

{\small\begin{verbatim}
You are a clarification assistant.
Your task: decide if the user's last
question is ambiguous or clear, using
only the conversation history and the
retrieved documents provided.

GROUNDING:
- Use ONLY information present in the
  conversation history or retrieved docs.
- Do NOT invent entities or details that
  are not in the context.

OUTPUT LOGIC:
- If the last user question has a clear
  interpretation that can be answered
  unambiguously, output exactly: CLEAR
- Otherwise, produce one clarifying
  question, following:
  (a) If there are 2-3 plausible options:
      list EACH with a short, factual
      description, then ask the user
      which one applies.
  (b) If there are 4-8 options: list them
      succinctly, then ask which one
      applies.
  (c) If there are >8 options OR no
      concrete options can be enumerated
      from context: ask an open-ended
      question (e.g., "Which <entity>
      are you referring to?").

STYLE:
- Be concise, neutral, and grounded.
- If listing options, they must be
  supported by the retrieved docs or
  conversation history.
\end{verbatim}}

\paragraph{User message.} The user message is constructed from each test record and
includes the conversation history, retrieved documents (if any), and the user's last
question, with placeholder strings when a component is empty:

{\small\begin{verbatim}
CONVERSATION HISTORY:
<prior user/assistant turns, or "(none)">

RETRIEVED DOCUMENTS:
<top-K retrieved documents, or "(no
 retrieved documents)">

LAST USER QUESTION:
<the user's last message>

Respond with either a grounded
clarifying question, or exactly CLEAR.
\end{verbatim}}

\section{Prompts for Generation Quality Metrics}
\label{sec:metric-prompts}

For the entity-option-centric metrics defined in Section~\ref{sec:eval-metrics}, an option is considered correct if it appears in the gold reference, is grounded in the retrieved documents, or is consistent with general knowledge and user intent. To compute precision and recall on free-form clarification responses, we need to implement judgments: (i) whether a clarification is open-ended, (ii) extracting the entity options it lists, (iii) checking whether an option is relevant to the ambiguous entity, and (iv) checking whether an option is covered by a reference list. We obtain these judgments with four LLM prompts.
All four prompts are executed with
Llama-3.3-70B-Instruct\footnote{\url{https://huggingface.co/meta-llama/Llama-3.3-70B-Instruct}}.
Placeholders in double braces (e.g.,~\texttt{\{\{ clarification question \}\}}) are
filled in at inference time.

\paragraph{Open-ended check.} Decides whether a clarification response is open-ended.

{\small\begin{verbatim}
[Instruction]
You are an assistant that determines
whether a clarification question is
open-ended. Output "yes" if it is
open-ended, "no" otherwise.

A clarification question is open-ended
(yes) if it allows for answers beyond
a fixed, limited set of options. A
clarification question is not
open-ended (no) if it restricts the
answer to a fixed and exhaustive set
of options.

Rules:
- If the question contains illustrative
  or example language such as "like",
  "such as", "for example",
  "including", or "include" anywhere in
  the question (including earlier
  sentences), treat the listed options
  as non-exhaustive and return yes,
  even if the final sentence asks
  "Which one are you referring to?".
- If the question lists choices plus a
  vague catch-all option (e.g.,
  "other", "else", "another"), return
  yes.
- If the question does not specify any
  concrete options and only contains a
  general clarification such as "Which
  [entity] do you mean?", return yes.
- If the question explicitly restricts
  the answer to a closed set of
  concrete options using language like
  "are you talking about A, B, or C",
  return no.

Clarification question:
{{ clarification question }}
\end{verbatim}}

\paragraph{Entity-option extraction.} Extracts explicit entity options from the
clarification response, separating concrete instances from catch-all choices.

{\small\begin{verbatim}
[Instruction]
You are an assistant that extracts
explicit options from clarification
questions. Output concrete entity
options into "instances" list and
catch-all options into "extra" list.

Rules and steps:
1. Identify all explicit options listed
   in the clarification question.
   - Put concrete entity names (e.g.,
     "UK", "Prime Minister of
     Australia") into "instances".
   - Put catch-all choices (e.g.,
     "else", "other", "another one")
     into "extra".
2. If an option has a description,
   extract only the main entity name,
   not the explanation.
3. Ignore numbering, bullet points, or
   letters (e.g., 1), 2., a), b)), and
   do not include them in the output.
4. If the question contains no explicit
   options, return "instances": [] and
   "extra": [].
5. Do not infer or add options that are
   not explicitly written in the text.

Clarification question:
{{ clarification question }}
\end{verbatim}}

\paragraph{Relevance check.} Checks whether an extracted option is a valid instance
of the ambiguous entity in the user query.

{\small\begin{verbatim}
[Instruction]
You are a relevance checker. Your task
is to determine whether a given check
choice is relevant to the ambiguous
entity in a user question. Output
"yes" if it is relevant, "no"
otherwise.

Rules and steps:
1. Identify the ambiguous entity type
   in the user question (e.g.,
   "ocean", "actor", "city").
2. Decide if the check choice is
   directly and literally a valid
   instance of that entity type.
   - The choice must exactly belong to
     the same entity type as the
     ambiguous entity, not just be
     related or associated with it.
   - Do not infer substitutions (e.g.,
     a character is not the same as an
     actor; a water mass is not the
     same as an ocean).
   - If the choice is related but not
     the same type, return no.
3. If the choice correctly specifies or
   narrows down the ambiguous entity,
   return yes.

User question:
{{ user question }}

Check choice:
{{ check choice }}
\end{verbatim}}

\paragraph{Coverage check.} Checks whether one option is covered by (or
equivalent to) any option in a reference list.

{\small\begin{verbatim}
[Instruction]
You are a coverage checker. Your task
is to determine if a check choice is
explicitly covered by or equivalent to
any choice in a given list of compared
choices. Output "yes" if it is
covered, "no" otherwise.

Rules:
1. If the check choice matches exactly
   (or is equivalent to) one of the
   compared choices, return yes.
   Otherwise, return no.
2. Equivalence of two choices:
   - They must refer to the same entity
     itself (not something merely
     related).
   - Valid equivalence cases:
     - Exact matches (e.g., "UK" vs
       "UK").
     - Substrings, abbreviations, or
       aliases of the same entity
       (e.g., "UK" vs
       "United Kingdom").
     - A country name and a role
       explicitly tied to that country
       are equivalent. Example:
       "Canada" == "Prime Minister of
       Canada".
   - Invalid equivalence cases:
     - Different entity types where
       one is not a country-role
       relationship (e.g., character
       vs actor, sea current vs
       ocean).

Check choice:
{{ check choice }}

Compared choices:
{{ compared choices }}
\end{verbatim}}

\section{Examples of Generation Quality Metrics}
\label{sec:eval-examples-appendix}

Table~\ref{tab:eval-examples} provides two worked examples that illustrate the
entity-option-centric precision, recall, and F1 metrics defined in
Section~\ref{sec:eval-metrics}. Example~1 shows a \emph{hedging over list}
clarification in which both the gold and generated clarifications enumerate specific
options. Example~2 shows an \emph{open-domain} clarification in which the gold
reference poses an open-ended question rather than listing specific options. For
such open-domain cases, we introduce an $[\textit{extra}]$ token to represent the
open-ended nature of the question: any correct generated option counts toward
precision, and the $[\textit{extra}]$ token in the gold is considered covered as long
as the generated clarification also includes an open-ended component.

\begin{table*}[t]
  \centering
  \small
  \begin{tabular}{p{2.3cm}p{6.1cm}p{6.1cm}}
    \toprule
     & \textbf{Example 1: Hedging over list} & \textbf{Example 2: Open-domain} \\
    \midrule
    \textbf{User query} & ``Is the language that descended from Common Brittonic still spoken today, or has it become extinct?'' & ``Where can I find information on the park amenities available at the New York county?'' \\[3pt]
    \textbf{Gold reference} & ``There are many languages that descended from Common Brittonic, such as: 1) Welsh, 2) Breton, and 3) Cornish. Which one are you referring to?'' & ``Which park are you interested in?'' \\[3pt]
    \textbf{Generated clarification} & ``There are several languages that descended from Common Brittonic, such as Welsh, Cornish, and Cumbric. Which one are you referring to?'' & ``There are many parks in New York county, such as Harriman State Park, Bear Mountain State Park, Minnewaska State Park, Rockland Lake State Park, and others. Which one are you interested in?'' \\[3pt]
    \textbf{Gold options} & \{Welsh, Breton, Cornish\} & \{$[\textit{extra}]$\} (open-ended, no specific options enumerated) \\[3pt]
    \textbf{Generated options} & \{Welsh, Cornish, Cumbric\} & \{Harriman State Park, Bear Mountain State Park, Minnewaska State Park, Rockland Lake State Park, $[\textit{extra}]$\} \\[3pt]
    \textbf{Precision} & 3/3 = 1.0 (Cumbric is a valid Brittonic language, thus correct) & 5/5 = 1.0 (all four parks are valid; open-ended component is appropriate) \\[3pt]
    \textbf{Recall} & 2/3 = 0.67 (Breton is missing from generated clarification) & 1/1 = 1.0 ($[\textit{extra}]$ covered by the open-ended question component) \\[3pt]
    \textbf{F1} & 0.80 & 1.0 \\
    \bottomrule
  \end{tabular}
  \caption{Examples of generation quality metric calculation.}
  \label{tab:eval-examples}
\end{table*}

\begin{table*}[t]
  \centering
  \small
  \begin{tabular}{lcccccccc}
    \toprule
    & \multicolumn{4}{c}{\emph{Classification Accuracy}} & \multicolumn{4}{c}{\emph{Generation Quality}} \\
    \cmidrule(lr){2-5} \cmidrule(lr){6-9}
    Training Configuration & Ambiguous & \shortstack{CLEAR-\\random} & \shortstack{CLEAR-\\hard} & Overall & Correct & Prec. & Rec. & F1 \\
    \midrule
    All training data      & 84.6\% & 98.5\% & 87.3\% & \textbf{92.4\%} & 88/104  & 91.7\% & 87.5\% & \textbf{89.5\%} \\
    Hard negatives removed & 97.1\% & 97.5\% & 35.4\% & 84.6\% & 101/104 & 88.1\% & 81.9\% & 84.9\% \\
    \bottomrule
  \end{tabular}
  \caption{Ablation study on Hard Negatives. The fine-tuned model is Granite-4.0-micro LoRA.}
  \label{tab:ablation}
\end{table*}

\section{Ablation Study on Hard Negatives}
\label{sec:ablation}

To isolate the contribution of hard negatives to the learned decision boundary, we
design an ablation experiment in which the only variable is the presence of
hard negatives in the training set. Specifically, we compare two
Granite-4.0-micro LoRA configurations: one trained on the full training set, and one 
trained on the same data but with all hard negatives
removed. Evaluation is performed on the full test set, enabling a direct measurement of 
how hard negatives influence both classification accuracy across the three categories 
and clarification generation quality.

Table~\ref{tab:ablation} shows that hard negatives are critical for the model's ability
to distinguish genuinely clear queries that look ambiguous. Without them, CLEAR-hard
accuracy collapses to 35.4\%, while Ambiguous detection
inflates to 97.1\%. With hard negatives, the model achieves a much better balance.
Generation quality also improves with hard negatives included in the training, likely
because the LoRA model learns more nuanced boundaries between ambiguous and clear queries,
leading to higher-quality clarifications when it does decide to clarify.

\begin{table*}[t]
  \centering
  \footnotesize
  \setlength{\tabcolsep}{4pt}
  \begin{tabular}{l cccc cccc}
    \toprule
    & \multicolumn{4}{c}{\emph{Banking (230)}} & \multicolumn{4}{c}{\emph{Telco (235)}} \\
    \cmidrule(lr){2-5} \cmidrule(lr){6-9}
    Model & Ambig. & C-hard & C-rand. & Overall & Ambig. & C-hard & C-rand. & Overall \\
    \midrule
    Granite-4.0-micro LoRA & 87.3\%          & 93.1\%          & 97.5\%          & 92.6\%          & 82.5\%          & 89.3\%          & 98.8\%          & 90.2\% \\
    \midrule
    GPT-4o                 & 53.2\%          & 61.1\%          & 96.2\%          & 70.4\%          & 48.8\%          & 68.0\%          & 93.8\%          & 70.2\%          \\
    GPT-5.6 Sol            & 86.1\%          & 29.2\%          & 67.1\%          & 61.7\%          & 82.5\%          & 38.7\%          & 67.5\%          & 63.4\%          \\
    GPT-OSS-120B           & 73.4\%          & 44.4\%          & 79.7\%          & 66.5\%          & 72.5\%          & 54.7\%          & 87.5\%          & 71.9\%          \\
    GPT-OSS-20B            & 81.0\%          & 37.5\%          & 63.3\%          & 61.3\%          & 80.0\%          & 46.7\%          & 72.5\%          & 66.8\%          \\
    Granite-4.1-30B        & 75.9\%          & 27.8\%          & 74.7\%          & 60.4\%          & 71.3\%          & 49.3\%          & 85.0\%          & 68.9\%          \\
    Granite-4.1-3B         & 22.8\%          & 80.6\%          & 86.1\%          & 62.6\%          & 40.0\%          & 82.7\%          & 83.8\%          & 68.5\%          \\
    Granite-4.0-micro      & 7.6\%           & 77.8\%          & 81.0\%          & 54.8\%          & 10.0\%          & 74.7\%          & 80.0\%          & 54.5\%          \\
    \bottomrule
  \end{tabular}
  \caption{Out-of-domain classification accuracy.}
  \label{tab:ood-classification}
\end{table*}

\begin{table*}[t]
  \centering
  \footnotesize
  \setlength{\tabcolsep}{4pt}
  \begin{tabular}{l cccc cccc}
    \toprule
    & \multicolumn{4}{c}{\emph{Banking (79 Ambiguous)}} & \multicolumn{4}{c}{\emph{Telco (80 Ambiguous)}} \\
    \cmidrule(lr){2-5} \cmidrule(lr){6-9}
    Model & Corr. & Prec. & Rec. & F1 & Corr. & Prec. & Rec. & F1 \\
    \midrule
    Granite-4.0-micro LoRA & 69/79 & 83.81\% & 79.42\% & 81.56\% & 66/80 & 95.03\% & 84.19\% & 89.28\% \\
    \midrule
    GPT-4o                 & 42/79 & 89.48\% & 90.16\% & 89.82\% & 39/80 & 86.67\% & 89.32\% & 87.97\% \\
    GPT-5.6 Sol            & 68/79 & 87.75\% & 67.06\% & 76.02\% & 66/80 & 93.52\% & 75.51\% & 83.55\% \\
    GPT-OSS-120B           & 58/79 & 84.99\% & 71.90\% & 77.90\%          & 58/80 & 87.64\% & 78.99\% & 83.09\%          \\
    GPT-OSS-20B            & 63/79 & 83.20\% & 67.41\% & 74.48\%          & 64/80 & 86.46\% & 71.74\% & 78.42\%          \\
    Granite-4.1-30B        & 60/79 & 92.88\% & 81.14\% & 86.61\%          & 57/80 & 79.90\% & 82.28\% & 81.07\%          \\
    Granite-4.1-3B         & 18/79 & 80.37\% & 66.67\% & 72.88\%          & 31/80 & 78.87\% & 69.09\% & 73.65\%          \\
    Granite-4.0-micro      & 6/79  & 66.67\% & 23.61\% & 34.87\%          & 5/80  & 70.00\% & 66.67\% & 68.29\%          \\
    \bottomrule
  \end{tabular}
  \caption{Out-of-domain generation quality on examples correctly classified
    as ambiguous.}
  \label{tab:ood-generation}
\end{table*}

\section{Out-of-Domain Evaluation Details}
\label{sec:ood-appendix}

The ClariEdge training and test data span four domains inherited from MTRAG
(ClapNQ, Cloud, FiQA, Govt). To assess how the fine-tuned LoRAs can
transfer to previously unseen domains, we construct an additional evaluation
set in two new domains, Banking and Telco, drawn from
MTRAG-UN~\citep{rosenthal2026mtragun}. MTRAG-UN provides 79 Banking and 80
Telco conversations, which we retain as-is for the CLEAR-random category
(their metadata already indicates that the queries are clear). We then
generate CLEAR-hard and Ambiguous examples in each domain following the
data-creation process described in Section~\ref{sec:data-creation}. The
Banking evaluation set contains 230 examples (79 Ambiguous, 72 CLEAR-hard,
and 79 CLEAR-random), of which 120 are pre-retrieval and 110 are
post-retrieval. The Telco evaluation set contains 235 examples (80 Ambiguous,
75 CLEAR-hard, and 80 CLEAR-random), of which 123 are pre-retrieval and 112
are post-retrieval. We evaluate the Granite-4.0-micro LoRA against the same
prompted baselines used in the in-domain evaluation.

\paragraph{Classification results.}
Table~\ref{tab:ood-classification} reports classification accuracy on the two
OOD domains. The Granite-4.0-micro LoRA leads on overall accuracy in both
domains, reaching 92.6\% on Banking and 90.2\% on Telco, close to its
92.4\% in-domain accuracy on the ClariEdge test set
(Table~\ref{tab:classification}). The class-imbalance pattern of prompted models
observed in-domain persists here: the larger prompted models (GPT-5.6 Sol, GPT-OSS-120B,
GPT-OSS-20B, and Granite-4.1-30B) over-clarify (high Ambiguous, low CLEAR-hard),
whereas the smaller prompted Granite models (Granite-4.1-3B and Granite-4.0-micro) under-clarify. Only the LoRA maintains balance across the three
categories on both OOD domains, which suggests that its ability to distinguish
genuine ambiguity from resolvable surface ambiguity transfers to new domains
rather than being overfit to the four training domains.

\paragraph{Generation quality results.}
Table~\ref{tab:ood-generation} reports the entity-option-centric generation
quality on the subset of examples where each model correctly predicted
Ambiguous. The Granite-4.0-micro LoRA achieves the highest F1 on Telco
(89.28\%). On Banking, prompted GPT-4o (89.82\%) and Granite-4.1-30B (86.61\%)
edge ahead of the LoRA in F1 (81.56\%), but on smaller correctly-flagged
subsets (42/79 and 60/79 vs.\ 69/79 for the LoRA), so their higher F1 comes
with substantially lower coverage.

Taken together, the OOD results indicate that the LoRA's advantage over
prompted baselines extends beyond the four training domains. The overall
picture (balanced classification across categories, high F1 on
correctly-flagged clarifications, and a large gap over prompted baselines
of similar size) matches the in-domain findings in
Sections~\ref{sec:cls-results} and~\ref{sec:gen-results}, suggesting that
the clarification skill acquired via ClariEdge is task-general rather than
domain-specific.

\end{document}